\def\paperlanguage{}
\pdfoutput=1 

\documentclass[letterpaper, 10 pt, conference]{ieeeconf}  

\usepackage{bm}
\usepackage{cite}
\usepackage{flushend}
\include{preamble}

\IEEEoverridecommandlockouts                              

\title{\LARGE \textbf
  {
    \switchlanguage%
    {%
      Online Learning of Danger Avoidance for Complex Structures of Musculoskeletal Humanoids and Its Applications
    }%
    {%
      筋骨格ヒューマノイドの複雑構造における危険回避のオンライン学習とその応用
    }%
  }
}

\author{Kento Kawaharazuka$^{1}$, Naoki Hiraoka$^{1}$, Yuya Koga$^{1}$, Manabu Nishiura$^{1}$, Yusuke Omura$^{1}$, Yuki Asano$^{1}$\\ Kei Okada$^{1}$, Koji Kawasaki$^{2}$, and Masayuki Inaba$^{1}$
  \thanks{$^{1}$ The authors are with the Department of Mechano-Informatics, Graduate School of Information Science and Technology, The University of Tokyo, 7-3-1 Hongo, Bunkyo-ku, Tokyo, 113-8656, Japan.
    {\texttt\small [kawaharazuka, hiraoka, koga, nishiura, omura, asano, k-okada, inaba]@jsk.t.u-tokyo.ac.jp}
  }
  \thanks{$^{2}$ The author is associated with TOYOTA MOTOR CORPORATION.
    {\texttt\small koji\_kawasaki@mail.toyota.co.jp}
  }
}
\begin{document}

\maketitle
\thispagestyle{empty}
\pagestyle{empty}

\begin{abstract}
  \switchlanguage%
  {%
    The complex structure of musculoskeletal humanoids makes it difficult to model them, and the inter-body interference and high internal muscle force are unavoidable.
    Although various safety mechanisms have been developed to solve this problem, it is important not only to deal with the dangers when they occur but also to prevent them from happening.
    In this study, we propose a method to learn a network outputting danger probability corresponding to the muscle length online so that the robot can gradually prevent dangers from occurring.
    Applications of this network for control are also described.
    The method is applied to the musculoskeletal humanoid, Musashi, and its effectiveness is verified.
  }%
  {%
    筋骨格ヒューマノイドは複雑な構造ゆえにモデル化が難しく, 身体の干渉や高い筋内力は避けられない.
    この問題を解決するために様々な安全機構が開発されているが, 危険が起こった時にそれを対処するだけでなく, 未然に危険を防ぐことも重要である.
    本研究では, 筋長指令に対応する危険度を表したネットワークをオンラインで学習し, ロボットが徐々に未然に危険を防ぐことができるようになる手法を提案する.
    また, このネットワークを用いた制御への応用についても述べる.
    本手法を筋骨格ヒューマノイドMusashiに適用し, その利点を示す.
  }%
\end{abstract}

\section{INTRODUCTION}\label{sec:introduction}
\switchlanguage%
{%
  The musculoskeletal humanoid \cite{nakanishi2013design, wittmeier2013toward, jantsch2013anthrob, asano2016kengoro} has various biomimetic advantages such as variable stiffness using redundant muscles, spherical joints without singular points, underactuated and flexible fingers, etc.
  At the same time, its complex musculoskeletal structure is difficult to model and various learning control methods have been developed \cite{ookubo2015learning, kawaharazuka2018online, kawaharazuka2019longtime, kawaharazuka2020autoencoder}.
  However, these methods mainly focus on the acquisition of static intersensory relationships, and there is always some error due to the influence of friction and hysteresis of muscles in the musculoskeletal structure.
  In addition, these methods only learn the relationship between joints and muscles, and the positional relationships and interferences between body links can only be obtained by using a geometric model.
  On the other hand, the ordinary self-collision avoidance cannot solve the problem, since the musculoskeletal structure has additional interferences among muscle wires, elastic elements, and various body parts.
  As a result, large internal muscle force may be generated unintentionally from wrong antagonistic relationships, or large muscle tension may be generated by inter-body interference due to the complex musculoskeletal structure and the discrepancy between the geometric model and the actual robot.

  When assuming the muscle length-based control, various safety mechanisms have been developed in order to solve this problem.
  In \cite{kawaharazuka2019longtime}, a safety mechanism to relax muscle length by taking temperature and muscle tension into account has been adopted.
  In \cite{kawaharazuka2019relax}, the internal muscle force is reduced by loosening muscles in turn, starting with unnecessary muscles so as not to change the current posture.
  \cite{nakashima2019healing} is a different approach, but it adopts a mechanism in which a self-healing module attached to the muscle breaks when a large force is applied, and returns to the original state after some time.
  These approaches work only when a large muscle tension is generated, that is, when a danger occurs, and they do not prevent it.
  Even if we elongate the muscle length in proportion to the muscle tension, it is not possible to quickly respond to the inter-body interference and large internal muscle force because they occur suddenly.

  In this study, we propose a method that enables the robot to learn the danger probability online and gradually prevent the danger from occurring.
  A neural network is used to represent the danger probability corresponding to the target muscle length, the control command of musculoskeletal humanoids.
  This network can be used to modify the target muscle length to a safe one and for prioritized inverse kinematics, which enable the robot to work while avoiding dangers.
  This study is organized as follows.
  In \secref{sec:musculoskeletal-humanoids}, basic characteristics of musculoskeletal structures and problems such as internal muscle force and inter-body interferences will be described.
  In \secref{sec:proposed}, we will describe the network structure, safety mechanism, initial training, online learning, and its applications.
  In \secref{sec:experiments}, we will conduct experiments on online learning of the network, modification of the target muscle length, and its use in prioritized inverse kinematics, respectively, to confirm its effectiveness.
  Finally, the discussion and conclusions will be presented.
}%
{%
  筋骨格ヒューマノイド\cite{nakanishi2013design, wittmeier2013toward, jantsch2013anthrob, asano2016kengoro}は冗長な筋肉による可変剛性・特異点のない球関節・劣駆動で柔軟な指等様々な生物模倣型の利点を有する.
  同時に, その複雑な筋骨格構造はモデル化が難しく, これまで様々な学習型制御手法が開発されてきている\cite{ookubo2015learning, kawaharazuka2018online, kawaharazuka2019longtime, kawaharazuka2020autoencoder}.
  しかしこれらの手法は主に静的なセンサ間関係のみの獲得に着目しており, 筋骨格構造に含まれる筋の摩擦やヒステリシスの影響から, 必ず多少の誤差が残る.
  また, これらの手法は関節・筋の関係を学習するのみであり, 骨格間の位置関係や干渉等は幾何モデルを用いる他ない.
  一方で, 筋骨格構造に置いては, さらに弾性バネや筋, 骨格らの間の干渉等の影響も有り, 通常の自己干渉回避では解決することができない.
  そのため, 間違った拮抗関係から意図せず大きな筋内力が発生してしまったり, 幾何モデルと実機のズレから身体同士が干渉して大きな筋張力が発生することがある.

  これまで, この問題を解決するために, 様々な安全機構が開発されてきた.
  \cite{kawaharazuka2019longtime}では温度と筋張力を考慮して筋長を緩める安全機構が採用されている.
  \cite{kawaharazuka2019relax}では姿勢を崩さないように必要のない筋から順番に筋を緩めていくことで, 筋内力を削減している.
  \cite{nakashima2019healing}は異なるアプローチであるが, 筋に装着した修復モジュールが大きな力が加わると断裂し, 時間が経つと元に戻るような機構を採用している.
  これらのアプローチは, 大きな筋張力が発生, つまり危険が起きたときに始めて作動し, それを未然に防ぐものではない.
  筋張力に比例して筋を伸ばすような動作をしたとしても, 干渉や大きな筋内力は突発的に起こるため, それに素早く対応することはできない.

  そこで本研究では, 制御指令に対応した危険度をオンラインで学習し, ロボットが徐々に未然に危険を防ぐことができるようになる手法を提案する.
  制御指令である筋長に対応した危険度をニューラルネットワークにより表現する.
  これを用いて指令筋長を安全なものに修正したり, 優先度付き逆運動学に使用することで危険を回避しながら動作させることが可能になる.
  本研究の構成は以下である.
  \secref{sec:musculoskeletal-humanoids}では基本的な筋骨格構造の特徴と, 筋内力や干渉等の問題点について述べる.
  \secref{sec:proposed}では提案手法におけるネットワーク構成・安全機構・初期学習・オンライン学習・その応用について述べる.
  \secref{sec:experiments}ではオンライン学習・筋長指令修正・優先度付き逆運動学での利用についてそれぞれ実験を行い, その有効性を確認する.
  最後に議論と結論を述べる.
}%

\begin{figure}[t]
  \centering
  \includegraphics[width=0.8\columnwidth]{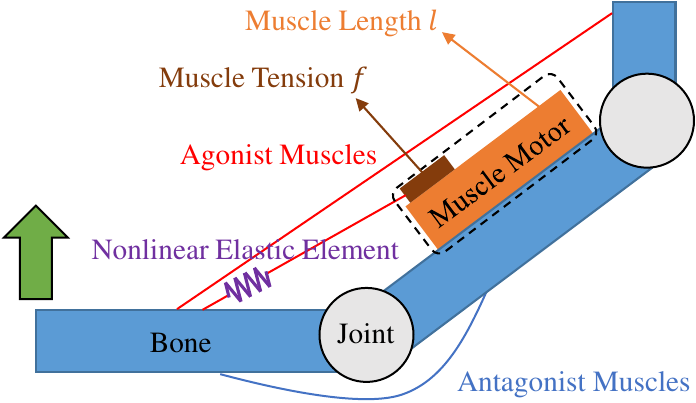}
  \caption{The basic structure of musculoskeletal humanoids.}
  \label{figure:musculoskeletal-structure}
  \vspace{-3.0ex}
\end{figure}

\section{Musculoskeletal Humanoids and Problems of the Complex Structure} \label{sec:musculoskeletal-humanoids}
\subsection{The Basic Structure of Musculoskeletal Humanoids} \label{subsec:basic-structure}
\switchlanguage%
{%
  The basic musculoskeletal structure is shown in \figref{figure:musculoskeletal-structure}.
  The redundant muscles are arranged antagonistically around the joints.
  The muscles are mainly composed of Dyneema, which is a synthetic fiber with high resistance to friction, and nonlinear elastic units that enable variable stiffness.
  In some robots, a movable pulley is used to fold back the muscle to increase the momentum arm.
  For soft environmental contact, the muscles are sometimes wrapped with soft foam cover as an exterior material, which makes the modeling more difficult.
  Muscle length $l$, muscle tension $f$, and muscle temperature $c$ can be measured for each muscle.
  The joint angle $\bm{\theta}$ cannot be usually measured due to the spherical joints and complex scapula; however, it can be measured with some robots, such as \cite{urata2006sensor, kawaharazuka2019musashi}.
  This study does not assume the measurement of joint angles.
}%
{%
  基本的な筋骨格構造を\figref{figure:musculoskeletal-structure}に示す.
  冗長な筋が関節の周りに拮抗して配置されている.
  筋は主に摩擦に強い合成繊維であるDyneemaによって構成されており, 可変剛性を可能とする非線形性弾性要素が筋と直列に配置されている場合が多い.
  ロボットによっては, モーメントアームを稼ぐために動滑車を使って筋を折り返している場合も存在する.
  柔軟な接触のために, 筋の周りには外装としての発泡材が巻かれている場合もあり, よりモデル化は困難を極める.
  それぞれの筋について筋長$l$・筋張力$f$・筋温度$c$が測定できる.
  関節角度$\bm{\theta}$は球関節や複雑な肩甲骨ゆえに測定できない場合が多いが, 一部のロボットで測定することが可能である\cite{urata2006sensor, kawaharazuka2019musashi}.
}%

\subsection{Problems of the Complex Structure} \label{subsec:complex-structure}
\begin{figure}[t]
  \centering
  \includegraphics[width=1.0\columnwidth]{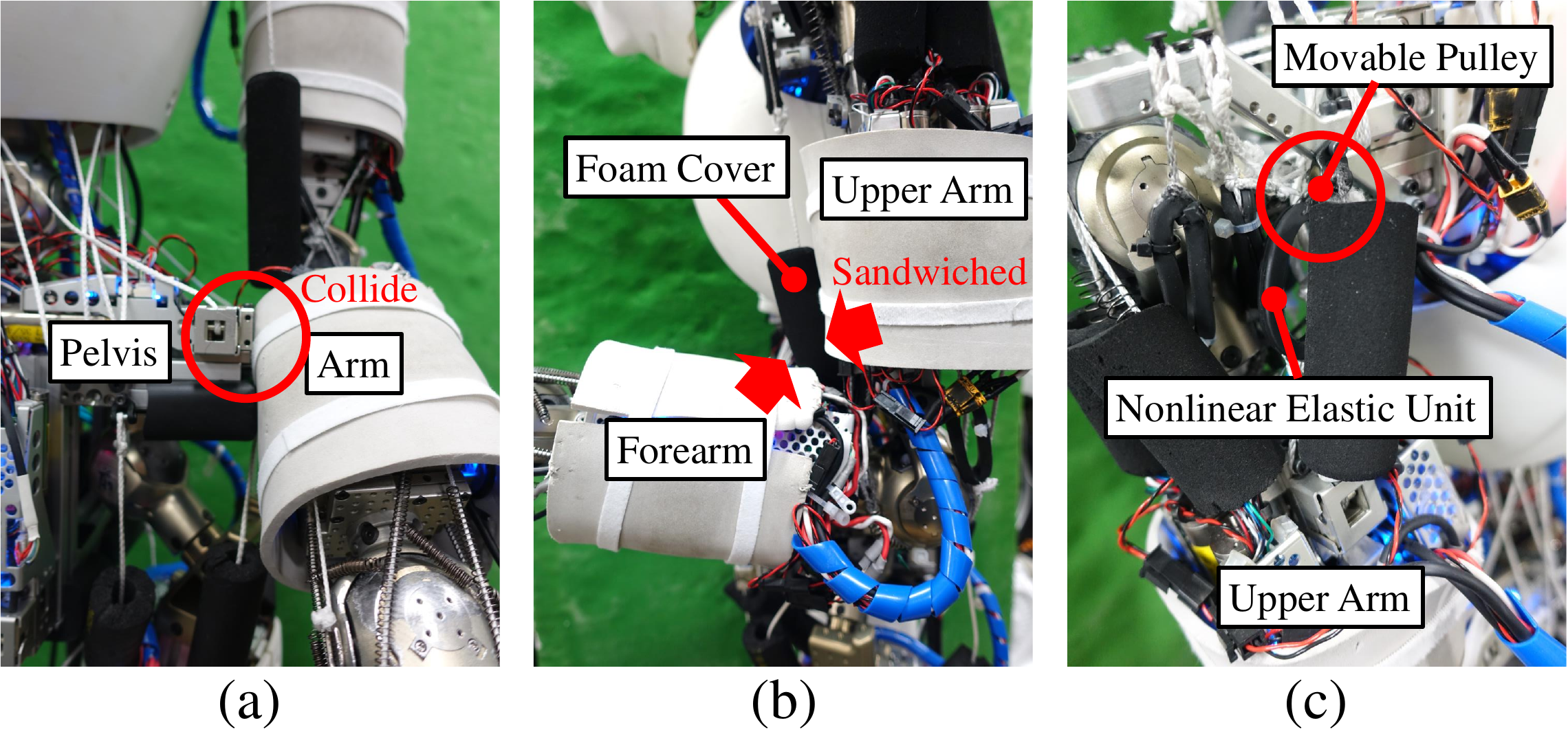}
  \caption{Various danger situations of musculoskeletal humanoids.}
  \label{figure:musculoskeletal-problem}
  \vspace{-3.0ex}
\end{figure}

\switchlanguage%
{%
  In this complex musculoskeletal structure, two main dangers can occur.
  First, the antagonistic relationship described in \secref{subsec:basic-structure} creates a closed-link structure, and large internal muscle force may be generated due to model errors.
  Second, inter-body interference may cause large muscle tension.
  Examples of this problem are shown in \figref{figure:musculoskeletal-problem}.
  Although it is possible to calculate the interference in advance from geometric models for ordinary axis-driven humanoids, this is not easy for musculoskeletal humanoids.
  First of all, not joint angle but muscle length is used as the control command, and when the target muscle length is sent, it is not always possible to realize the intended posture completely due to model errors.
  In addition, due to the complex structure, there are not only cases where the waist and the forearm link simply interfere as in (a) of \figref{figure:musculoskeletal-problem}, but also cases where a soft foam cover is sandwiched between the upper arm and forearm links and interferes with them as in (b).
  In addition, there is a situation in which a movable pulley interferes with a nonlinear elastic unit and the muscle cannot be pulled any further, as in (c).
  It is desirable to update the probability of these dangerous situations at all times because they are constantly changing due to aging and other factors.
}%
{%
  この複雑な筋骨格構造では主に2つの危険状態が起こりえる.
  １つ目に, \secref{subsec:basic-structure}で述べた拮抗関係は閉リンク構造を作るため, モデル誤差により大きな筋内力が発生する可能性がある.
  ２つ目に, 骨格間の干渉により, 大きな筋張力が発生する可能性がある.
  その例を\figref{figure:musculoskeletal-problem}に示す.
  通常の軸駆動ヒューマノイドであれば幾何モデルから事前に干渉を計算することが可能であるが, 筋骨格構造においてはそうは行かない.
  まず, 制御指令が関節角度ではなく筋長であり, その指令筋長を送ったとしても, 学習誤差等によって完全にその関節角度を実現できるとは限らない.
  また, 複雑な構造ゆえに, \figref{figure:musculoskeletal-problem}の(a)のように単純に腰リンクと前腕リンクが干渉する場合だけでなく, (b)のように上腕リンクと前腕リンクの間に発泡性の筋外装が挟まれて干渉する場合もある.
  この他, (c)のように筋の折り返し部品と非線形弾性要素が干渉し, それ以上筋を引っ張ることができなる状態も存在する.
  そしてこの危険状態は経年劣化等によって常に変化していくものであり, 逐次的に更新していくことが望ましい.
}%

\begin{figure*}[t]
  \centering
  \includegraphics[width=1.8\columnwidth]{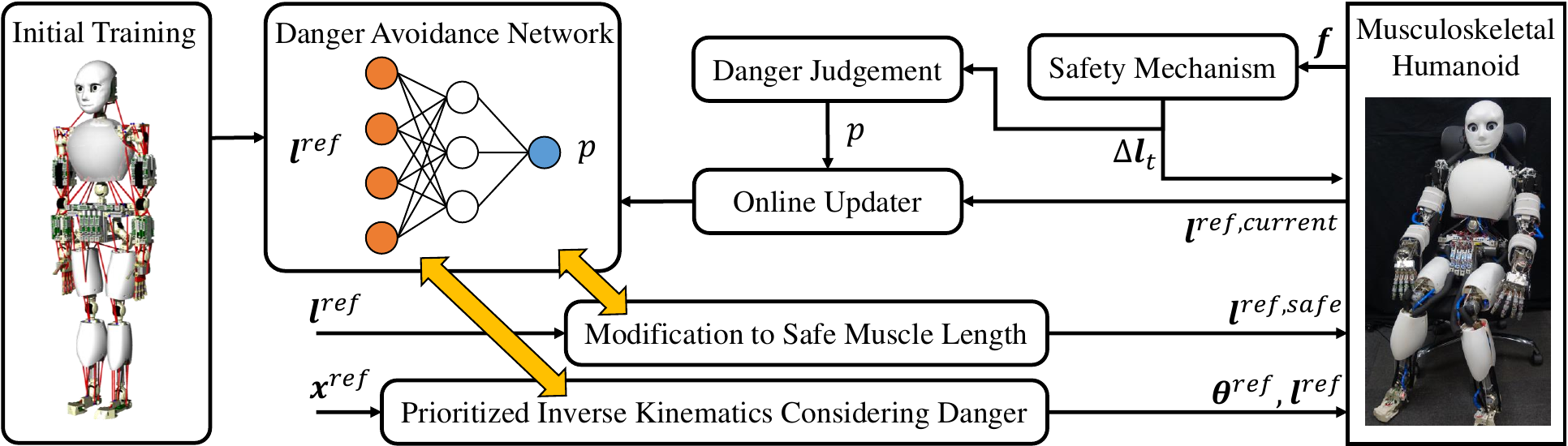}
  \caption{The whole system of danger avoidance for musculoskeletal humanoids.}
  \label{figure:whole-system}
  \vspace{-3.0ex}
\end{figure*}

\section{Online Learning of Danger Avoidance} \label{sec:proposed}
\switchlanguage%
{%
  A complete picture of the danger avoidance system is shown in \figref{figure:whole-system}.
}%
{%
  危険回避システムの全体像を\figref{figure:whole-system}に示す.
}%

\subsection{Network Structure} \label{subsec:network-structure}
\switchlanguage%
{%
  The structure of Danger Avoidance Network (DAN) proposed in this study is described.
  For axis-driven robots, joint angle $\bm{\theta}^{ref}$ is used as the control command, whereas for musculoskeletal robots, muscle length $\bm{l}^{ref}$ is used as the control command ($\{\bm{\theta}, \bm{l}\}^{ref}$ represent the target values of $\{\bm{\theta}, \bm{l}\}$).
  Therefore, while self-collision probability corresponding to $\bm{\theta}^{ref}$ can be obtained in the axis-driven type, the probability of dangers corresponding to $\bm{l}^{ref}$ can be obtained in the musculoskeletal type.
  The dangers here include high internal muscle force due to errors in the antagonistic relationship and high muscle tension due to inter-body interference described in \secref{subsec:complex-structure}.
  If the danger corresponding to $\bm{\theta}^{ref}$ is expressed in the musculoskeletal type, it is not possible to take into account the internal muscle force and inter-body interferences that vary with the state of muscles.
  Therefore, in this study, we train the following function $h_{dan}$,
  \begin{align}
    p = h_{dan}(\bm{l}^{ref})
  \end{align}
  where $p$ denotes the danger probability ($0\leq{p}\leq1$), whose definition will be described subsequently.
  $\bm{l}^{ref}$ is a $m$-dimensional vector ($m$ represents the number of related muscles).

  We represent the function $h_{dan}$ by a neural network.
  In this study, the neural network is composed of four fully-connected layers, and the number of units of each layer is set to $m$, 64, 64, and 1.
  Batch Normalization \cite{ioffe2015batchnorm} is applied after each layer except for the last layer.
  The activation function of the middle layer is ReLU \cite{nair2010relu}, and that of the last layer is Sigmoid to output probability from 0 to 1.
}%
{%
  本研究で提案するDanger Avoidance Network (DAN)の構造について説明する.
  軸駆動型ロボットは制御入力が関節角度$\bm{\theta}^{ref}$なのに対し, 筋骨格型ロボットは制御入力が筋長$\bm{l}^{ref}$となる(ここで, $\{\bm{\theta}, \bm{l}\}^{ref}$は$\{\bm{\theta}, \bm{l}\}$の指令値を表す).
  そのため, 軸駆動型では$\bm{\theta}^{ref}$に対応したリンク間干渉が得られるのに対して, 筋骨格型では$\bm{l}^{ref}$に対応した危険度が得られる.
  ここで言う危険には, \secref{subsec:complex-structure}で述べた拮抗関係の誤差による高い筋内力と干渉による高い筋張力が含まれる.
  もし筋骨格型において$\bm{\theta}^{ref}$に対応する危険度を表すとすると, 筋内力や, \figref{figure:musculoskeletal-problem}にあるような筋の状態によって変化する干渉を考慮することができない.
  よって, 本研究では以下の関数$h_{dan}$を学習していく.
  \begin{align}
    p = h_{dan}(\bm{l}^{ref})
  \end{align}
  ここで, $p$は危険度($0\leq{p}\leq1$)を表し, その定義については後に説明する.
  また, $\bm{l}^{ref}$は$m$次元のベクトルである($m$は関係する筋の数を表す).

  この関数$h_{dan}$をニューラルネットワークで表現する.
  本研究では4層のニューラルネットワークとし, それぞれのユニット数は$m$, 64, 64, 1とした.
  最終層以外の各層の後にはBatch Normalization \cite{ioffe2015batchnorm}を適用した.
  中間層の活性化関数はReLU \cite{nair2010relu}であり, 最終層は0から1の確率を出すためSigmoidとした.
}%

\begin{figure}[t]
  \centering
  \includegraphics[width=1.0\columnwidth]{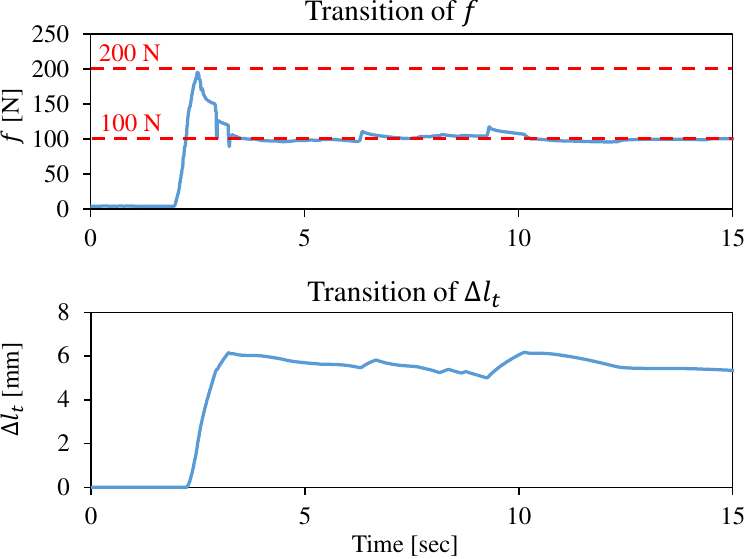}
  \caption{The transition of $f$ and $\Delta{l}_{t}$ when  contracting one muscle by 40 mm over 0.5 seconds with a safety mechanism.}
  \label{figure:safety-mechanism}
  \vspace{-3.0ex}
\end{figure}

\subsection{Safety Mechanism} \label{subsec:safety-mechanism}
\switchlanguage%
{%
  In this study, we provide not only danger avoidance, but also a safety mechanism to mitigate the danger at the same time.
  When the safety mechanism is activated, the current situation is considered to be dangerous, and the danger probability $p$ is defined.
  For each muscle, the safety mechanism calculates the degree of muscle relaxation $\Delta{l}_{t}$ according to muscle tension, and sends $l^{ref}+\Delta{l}_{t}$ to the actual robot as follows,
  \begin{align}
    &if\;\;f > f^{thre} \nonumber\\
    &\;\;\;\;\;\;\Delta{l}_{t} = \Delta{l}_{t-1} + \max(-C_{minus}d, \min(C_{gain}d-\Delta{l}_{t-1}, C_{plus}d))\nonumber\\
    &else \nonumber\\
    &\;\;\;\;\;\;\Delta{l}_{t} = \Delta{l}_{t-1} + \max(-C_{minus}d, \min(0-\Delta{l}_{t-1}, C_{plus}d))\nonumber\\
    &d = |f-f^{thre}|
  \end{align}
  where $f^{thre}$ is the threshold of muscle tension $f$ that begins to elongate the muscle length, $|\bm{\bullet}|$ is the absolute value, $\Delta{l}_{t}$ is the degree of relaxation at time step $t$, $C_{\{minus, plus\}}$ is a coefficient that determines the amount of muscle length change in one time step in the negative or positive direction, and $C_{gain}$ is a coefficient that determines the maximum amount of relaxation.
  In other words, the muscle is relaxed and tensed so that the muscle tension does not exceed the maximum value, while limiting the change in muscle length by $C_{minus}d$ and $C_{plus}d$.
  In this study, we set $f^{thre}=200$ [N], $C_{minus}=0.001$ [mm/N], $C_{plus}=0.003$ [mm/N], and $C_{gain}=2.0$ [mm/N], and this control works with a $8$ msec period.

  The behavior of the safety mechanism is shown in \figref{figure:safety-mechanism}, when the end of the muscle is fixed and the muscle is contracted by -40 mm over 0.5 s.
  In this experiment only, $f^{thre}=100$ [N] for safety reasons.
  The muscle tension is momentarily increased to 200 N and then suppressed to 100 N.
  Thus, although the safety mechanism lowers the increased muscle tension to $f^{thre}$ instantly, its peak cannot be suppressed.
  If there is no learning system of danger avoidance, the same dangerous behaviors will occur again and again.

  Finally, we define the danger probability $p$ as follows.
  \begin{align}
    p=\left\{
      \begin{array}{ll}
        1.0 & \Delta{l}_{t} > 0.0\\
        0.0 & otherwise \label{eq:p-definition}
      \end{array}
      \right.
  \end{align}
  In other words, when the safety mechanism is activated, we consider it to be a danger, and the rest of the time we consider it not to be a danger.
  This is only an example, and it is possible to define the behavior that the user feels is dangerous in a completely different way.
}%
{%
  本研究では危険回避だけでなく, 危険時にそれを緩和する安全機構も同時に備える.
  そして, この安全機構が作動したときを危険と見なし, 危険度$p$を定義する.
  それぞれの筋について, 安全機構は以下のように筋張力に応じて筋長弛緩度$\Delta{l}_{t}$を計算し, 実機には$l^{ref}+\Delta{l}_{t}$を送る.
  \begin{align}
    &if\;\;f > f^{thre} \nonumber\\
    &\;\;\;\;\;\;\;\;\;\;\;\;\Delta{l}_{t} = \Delta{l}_{t-1} + min(C_{gain}d-\Delta{l}_{t-1}, C_{plus}d)\nonumber\\
    &else \nonumber\\
    &\;\;\;\;\;\;\;\;\;\;\;\;\Delta{l}_{t} = \Delta{l}_{t-1} + max(0-\Delta{l}_{t-1}, -C_{minus}d)\nonumber\\
    &d = |f-f^{thre}|
  \end{align}
  ここで, $f^{thre}$は筋長を弛緩させ始める筋張力$f$の閾値, $|\bm{\bullet}|$は絶対値, $\Delta{l}_{t}$はタイムステップ$t$における弛緩度, $C_{\{minus, plus\}}$はマイナス方向またはプラス方向に対する一ステップの筋長変化量を決める係数, $C_{gain}$は最大弛緩量を決める係数である.
  つまり, $C_{minus}d$, $C_{plus}d$で制限をかけながら, 筋張力が最大値を越えないように筋を弛緩・緊張させている.
  本研究では, $f^{thre}=200$ [N], $C_{minus}=0.001$ [mm/N], $C_{plus}=0.003$ [mm/N], $C_{gain}=2.0$ [mm/N]とし, 本制御は$8$ msec周期で行う.

  一本の筋を張った状態で末端を固定し, 0.5秒で-40 mm緊張させたときの, 安全機構の挙動を\figref{figure:safety-mechanism}に示す.
  また本実験のみ, 安全のため$f^{thre}=100$ [N]としている.
  筋張力は一瞬だけ200 Nまで上がり, その後100 Nまで抑えられている.
  このように, 安全機構は高まった筋張力を瞬時に$f^{thre}$まで下げてくれるものの, そのピークは抑えられない
  また, 学習等のシステムがなければ, 再び同じように危険な動作を何度も行ってしまう.

  最後に, 以下のように危険度$p$を定義する.
  \begin{align}
    p=\left\{
      \begin{array}{ll}
        1.0 & \Delta{l}_{t} > 0.0\\
        0.0 & otherwise \label{eq:p-definition}
      \end{array}
      \right.
  \end{align}
  つまり, 安全機構が作動したとき, それを危険と見なし, それ以外は危険がないと見なす.
  これはあくまで一例に過ぎず, ユーザが危険と感じた動作を全く異なった形で定義することも可能である.
}%

\subsection{Initial Training} \label{subsec:initial-training}
\switchlanguage%
{%
  We perform the initial training of DAN.
  In this procedure, the actual robot is not necessary.
  Although we can start with a completely random state of DAN weights, we initialize the network using joint-muscle space mapping in order to achieve faster convergence in online learning.
  First, we determine the lower and upper limits of each joint angle on the geometric model, and then sample the joint angle values randomly over a slightly wider range than the lower and upper limits (in this study, $\pm10$ deg).
  If any joint angle deviates from the limits, it is assumed to be dangerous, i.e., $p=1.0$, and otherwise $p=0.0$.
  Using the mapping between joints and muscle space obtained in \cite{kawaharazuka2019longtime}, the target joint angle $\bm{\theta}^{ref}$ is converted to the target muscle length $\bm{l}^{ref}$ (target muscle tension in \cite{kawaharazuka2019longtime} is set to 10 N), and the data of $(\bm{l}^{ref}, p)$ is accumulated.
  This dataset is used to train DAN.
  We always add gaussian noise with a mean of 0 and a standard deviation of $C_{st}$ to $\bm{l}^{ref}$, to make the network more robust to noise.
  In this study, $C_{st}=3$ [mm], the number of data is 12000, the number of batches is 100, the number of epochs is 100, and the update rule is Adam \cite{kingma2015adam}.
}%
{%
  DANの初期学習を行う.
  DANの重みが全くランダムな状態から始めることもできるが, オンライン学習におけるより速い収束を目指し, 関節-筋空間マッピングを用いてネットワークを初期化する.
  まず幾何モデル上でそれぞれの関節角度の下限と上限を決め, その値よりも少し広い範囲(本研究では$\pm10$ deg)でランダムに関節角度の値を一様サンプリングする.
  このとき, 一つでもサンプリングされた値が下限上限を超えた関節があれば, それは危険, つまり$p=1.0$とし, そうでなければ$p=0.0$とする.
  \cite{kawaharazuka2019longtime}で得られた関節と筋空間のマッピングを用いてその関節角度指令$\bm{\theta}^{ref}$を筋長$\bm{l}^{ref}$に変換し(なお, 筋張力は一律10 Nとしている), $(\bm{l}^{ref}, p)$のデータを蓄積する.
  このデータセットを用いてDANを学習させる.
  なお, ノイズに強くするため, $\bm{l}^{ref}$には常に平均0, 標準偏差$C_{st}$のガウシアンノイズを加える.
  本研究では$C_{st}=3$ [mm]とし, データ数を12000, バッチ数を100, エポック数を100, 更新方法をAdam \cite{kingma2015adam}として学習させた.
}%

\subsection{Online Learning} \label{subsec:online-learning}
\switchlanguage%
{%
  We update DAN online using the actual robot sensor information.
  First, we obtain the current target muscle length $\bm{l}^{ref, current}$ sent to the actual robot.
  Then, we obtain the danger probability $p^{predicted}$ that is estimated when we feed it into the current network.
  At the same time, we also get $p$ from \equref{eq:p-definition}, and determine whether the current state of the actual robot is in danger or not.
  Here, if ($p=1.0$ and $p^{predicted}<0.9$) or ($p=0.0$ and $p^{predicted}>0.1$), then $(\bm{l}^{ref, current}, p)$ is accumulated as data.
  This corresponds to the accumulation of data when the predicted and actual danger states differ significantly (($p=1.0$ and $p^{predicted}<0.1$) or ($p=0.0$ and $p^{predicted}>0.9$)) and when the probability is ambiguous ($0.1 < p < 0.9$).
  Moreover, this accumulation is performed only when $||\bm{l}^{ref, current}-\bm{l}^{ref, pre}||_{2}>C_{diff}$ against the previously accumulated $\bm{l}^{ref, pre}$ ($||\bm{\bullet}||_{2}$ is L2 norm and $C_{diff}$ is the threshold of the difference of target muscle lengths).
  That is, data is not accumulated unless the robot moves to some extent.
  The maximum number of data $N_{max}$ is determined, and if more data is accumulated than $N_{max}$, the oldest data is deleted.
  When the number of data exceeds a threshold $N_{thre}$, DAN is updated online with all the accumulated data each time new data is obtained.

  In this study, we set $C_{diff}=20.0$ [mm], $N_{max}=100$, and $N_{thre}=30$.
  The number of batches and epochs for online learning is set to 10 and 3, respectively, and the update rule is Momentum SGD.
}%
{%
  DANを実機データを使ってオンラインに更新していく.
  まず, 実機に現在送られている指令筋長$\bm{l}^{ref, current}$を取得する.
  これを現在のネットワークに入れたときに推測された危険度$p^{predicted}$を取得する.
  また同時に, \equref{eq:p-definition}から$p$を取得し, 現在の実機状態が危険状態であるかどうかを判定する.
  ここで, $p=1.0 and p^{predicted}<0.9$または$p=0.0 and p^{predicted}>0.1$の場合に$(\bm{l}^{ref, current}, p)$をデータとして蓄積する.
  これは, 予測と実際の危険度が大きく違う場合($p=1.0 and p^{predicted}<0.1$または$p=0.0 and p^{predicted}>0.9$), また, 確率が曖昧である場合($0.1 < p < 0.9$)にデータを蓄積することに相当する.
  また, このデータ蓄積は, 一つ前に蓄積された$\bm{l}^{ref, pre}$に対して, $||\bm{l}^{ref, current}-\bm{l}^{ref, pre}||_{2}>C_{diff}$の場合にのみ実行される($||\bm{\bullet}||_{2}$はL2ノルム, $C_{diff}$は指令筋長差分の閾値とする).
  つまり, ロボットがある程度動かない限りはデータは蓄積されない.
  データの最大数$N_{max}$を決め, それより多いデータが蓄積された場合は最も古いデータを削除する.
  データ数が閾値$N_{thre}$を超えたところから, 新しいデータが得られる度に, 蓄積された全データを使ってオンラインでDANを更新する.

  本研究では, $C_{diff}=20.0$ [mm], $N_{max}=100$, $N_{thre}=30$とする.
  また, オンライン学習の際のバッチ数は10, エポック数は3とし, 更新則はMomentum SGDを用いる.
}%

\subsection{Applications} \label{subsec:applications}
\switchlanguage%
{%
  Using the obtained DAN, we can predict the danger probability $p^{predicted}$ before sending $\bm{l}^{ref}$ to the actual robot and decide whether to move it or not.
  In addition to such a way of use, two applications of DAN are described below.

  First, we show how to modify a certain target muscle length $\bm{l}^{ref}$ to a less dangerous target muscle length.
  When $p^{predicted}$ is obtained from $\bm{l}^{ref}$ by using DAN, if $p^{predicted}>0.1$, it is potentially dangerous, and we should not send $\bm{l}^{ref}$ as it is.
  Therefore, we propose to update the target muscle length using backpropagation technique \cite{rumelhart1986backprop} and gradient descent as follows to calculate a safe target muscle length $\bm{l}^{ref, safe}$ and send it to the actual robot,
  \begin{align}
    L(\bm{l}^{ref, safe}) &= |h_{dan}(\bm{l}^{ref, safe})|+C_{loss}||\bm{l}^{ref}-\bm{l}^{ref, safe}||_{2}\\
    \bm{l}^{ref, safe} &\gets \bm{l}^{ref, safe} - \gamma \partial L/\partial \bm{l}^{ref, safe}
  \end{align}
  where $C_{loss}$ is the weighting constant, $L$ is the loss, and $\gamma$ is the learning rate.
  Here, as the initial value of $\bm{l}^{ref, safe}$ to be updated, the current target muscle length $\bm{l}^{ref, current}$ is used.
  That is, $\bm{l}^{ref, safe}$ approaches the original target muscle length $\bm{l}^{ref}$ from the current target muscle length $\bm{l}^{ref, current}$, but when the danger probability $p^{predicted}$ increases, the update is stopped, and the safe target muscle length $\bm{l}^{ref, safe}$ before the danger state is obtained.
  Although $\gamma$ can be a fixed value, in this study, the maximum $\gamma^{max}$ and the number of batches $N_{batch}$ are determined, and $N_{batch}$ learning rates less than $\gamma^{max}$ is used as in line search methods.
  After updating $\bm{l}^{ref, safe}$ with each learning rate, we compute $L$ again and adopt the one with the smallest $L$.
  This process is repeated $N_{iter}$ times to compute $\bm{l}^{ref, safe}$.
  In this study, we set $C_{loss}=0.01$, ${\gamma}^{max}=0.1$, $N_{batch}=10$, and $N_{iter}=30$.

  Next, we consider the use of DAN in prioritized inverse kinematics \cite{kanoun2009prioritized}.
  We solve inverse kinematics for the target end effector coordinate $\bm{x}^{ref}$ and calculate the target joint angle $\bm{\theta}^{ref}$ and the target muscle length $\bm{l}^{ref}$.
  The first task is set as $\bm{x}^{ref} = h_{kinematics}(\bm{\theta}^{ref})$ ($h_{kinematics}$ denotes the function that converts the joint angle to the end effector coordinate).
  Then, the obtained $\bm{\theta}^{ref}$ is converted to muscle length by \cite{kawaharazuka2019longtime}, and the danger probability $p^{predicted}$ is predicted by DAN.
  When $p^{predicted}>0.1$, its posture $\bm{\theta}^{ref}$ is potentially dangerous, so we set it as an avoidance posture $\bm{\theta}^{avoid}$.
  Next, as a second task, we set $|\bm{\theta}^{ref}-\bm{\theta}^{avoid}|>d$ to solve the prioritized inverse kinematics \cite{kanoun2009prioritized} ($d$ represents a threshold constant, which is set to 0.2 rad in this study).
  Similarly, $p^{predicted}$ is predicted, and if the probability does not decrease below 0.1, the number of $\bm{\theta}^{avoid}$ will increase, and the constraints of the second task will increase.
  Actually, the first and second tasks are linearly approximated around the current target joint angle to be used in \cite{kanoun2009prioritized}.
}%
{%
  得られたDANを用いて, 実機に$\bm{l}^{ref}$を送る前に危険度$p^{predicted}$を予測し, 動作を行うかどうかを決定することができる.
  これだけでなく, 本章ではDANを用いた2つの応用について述べる.

  まず, ある指令筋長$\bm{l}^{ref}$を送る際に, これをなるべく危険ではない筋長指令へと修正する方法について述べる.
  $\bm{l}^{ref}$からDANを用いて$p^{predicted}$を求めた時, これが$p^{predicted}>0.1$であれば危険な可能性があるため, そのまま$\bm{l}^{ref}$を送るわけにはいかない.
  そこで, 以下のように指令筋長$\bm{l}^{ref, safe}$を更新することで, 安全な筋長指令$\bm{l}^{ref, safe}$を計算し, これを実機に送ることを考える.
  \begin{align}
    L(\bm{l}^{ref, safe}) &= |h_{dan}(\bm{l}^{ref, safe})|+C_{loss}||\bm{l}^{ref}-\bm{l}^{ref, safe}||_{2}\\
    \bm{l}^{ref, safe} &\gets \bm{l}^{ref, safe} - \gamma \partial L/\partial \bm{l}^{ref, safe}
  \end{align}
  ここで, $C_{loss}$は重み付けの定数, $L$は損失, $\gamma$は学習率とする.
  ここで, $\bm{l}^{ref, safe}$の初期値は現在の指令筋長である$\bm{l}^{ref, current}$を用いる.
  つまり, $\bm{l}^{ref, safe}$は現在の指令筋長$\bm{l}^{ref, current}$から送るべき指令筋長$\bm{l}^{ref}$に徐々に近づいていくが, 危険度$p^{predicted}$が高くなると途中で更新が止まり, 危険になる手前の指令筋長$\bm{l}^{ref, safe}$が得られることになる.
  $\gamma$はある固定の値でも良いが, 本研究では$\gamma$の最大値$\gamma^{max}$とバッチ数$N_{batch}$を決め, $\gamma^{max}$以下の$N_{batch}$個の学習率を用いる.
  それぞれの学習率で$\bm{l}^{ref, safe}$を更新したあともう一度$L$を計算し, 最も$L$が小さかったものを採用する.
  この処理を$N_{iter}$回繰り返すことで, $\bm{l}^{ref, safe}$を計算する.
  本研究では$C_{loss}=0.01$, ${\gamma}^{max}=0.1$, $N_{batch}=10$, $N_{iter}=30$とする.

  次に, 優先度付き逆運動学\cite{kanoun2009prioritized}にDANを利用することを考える.
  ある手先の姿勢$\bm{x}^{ref}$に対して逆運動学を解き, 指令関節角度$\bm{\theta}^{ref}$, そのときの指令筋長$\bm{l}^{ref}$を計算する.
  第一タスクとして$\bm{x}^{ref} = h_{kinematics}(\bm{\theta}^{ref})$を設定する($h_{kinematics}$は関節角度を手先位置に変換する関数を表す).
  このとき, 得られた$\bm{\theta}^{ref}$を\cite{kawaharazuka2019longtime}によって筋長に変換し, DANによって危険度$p^{predicted}$を予測する.
  $p^{predicted}>0.1$のとき, その姿勢$\bm{\theta}^{ref}$は危険な可能性があるため, 回避姿勢$\bm{\theta}^{avoid}$として設定する.
  次に, 第二タスクとして, $|\bm{\theta}^{ref}-\bm{\theta}^{avoid}|>d$を設定して, 優先度付き逆運動学\cite{kanoun2009prioritized}を解く($d$は閾値の定数を表す, 本研究では0.2 radとする).
  同様に$p^{predicted}$を予測し, 危険度が下がらなければこの$\bm{\theta}^{avoid}$は増えていき, 第二タスクの条件は増加していくことになる.
  また, 実際には第一タスク, 第二タスクはそれぞれ現在の関節角度の周りに線形近似され\cite{kanoun2009prioritized}において用いられる.
}%

\begin{figure}[t]
  \centering
  \includegraphics[width=1.0\columnwidth]{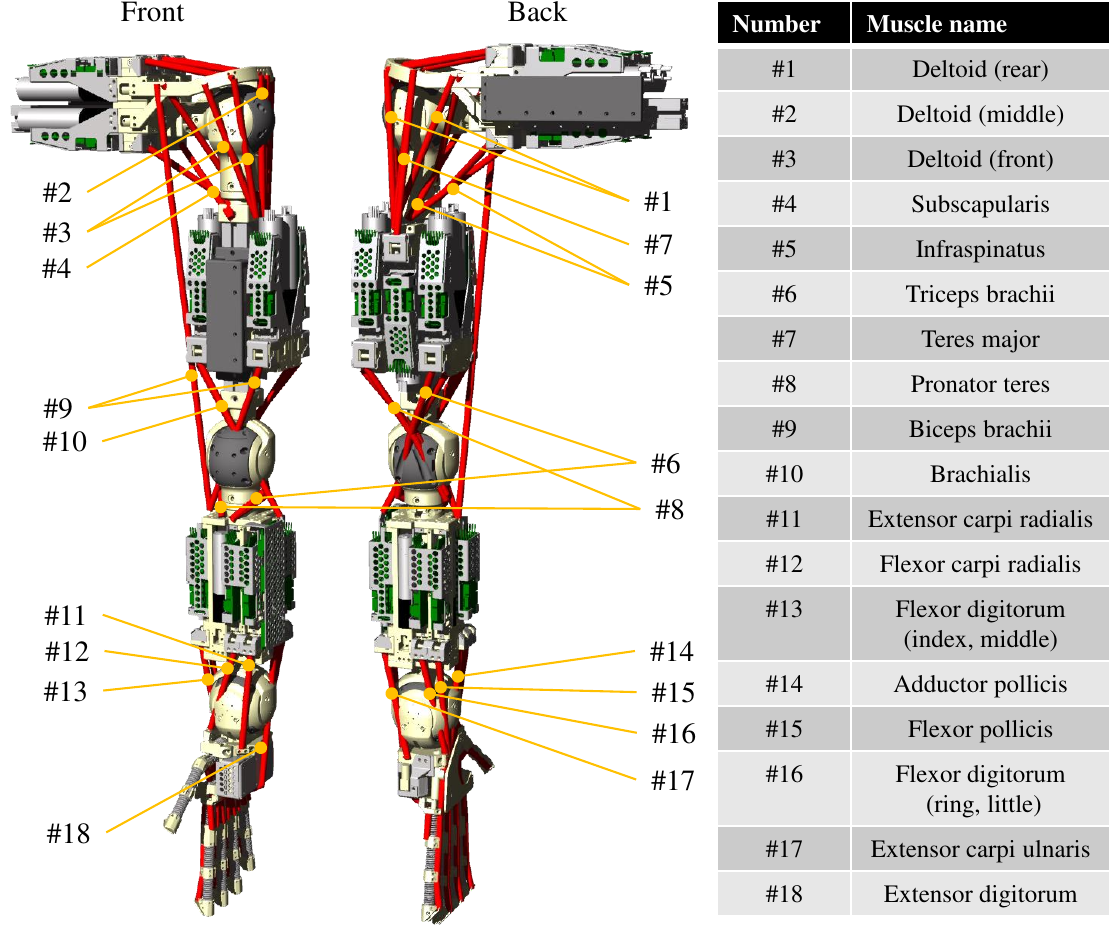}
  \caption{Muscle arrangement of the musculoskeletal humanoid Musashi \cite{kawaharazuka2019musashi} used in this study.}
  \label{figure:experimental-setup}
  \vspace{-1.0ex}
\end{figure}

\section{Experiments} \label{sec:experiments}
\subsection{Experimental Setup} \label{subsec:experimental-setup}
\switchlanguage%
{%
  In this study, the musculoskeletal humanoid Musashi \cite{kawaharazuka2019musashi} is used for experiments.
  Its muscle arrangement is shown in \figref{figure:experimental-setup}, which mimics the major muscles of the human body.
  A nonlinear elastic unit using Grommet is attached at the end of the muscle wire (Dyneema), and the muscle wire is surrounded by a spring and a soft foam cover (the pictures of \figref{figure:musculoskeletal-problem} are parts of Musashi).
  In this study, we mainly used five degrees of freedom, three for the shoulder and two for the elbow.
  There are 10 muscles related to these joints, and one of them is a polyarticular muscle.
}%
{%
  本研究では筋骨格ヒューマノイドMusashiを実験に用いる\cite{kawaharazuka2019musashi}.
  筋配置は\figref{figure:experimental-setup}のようになっており, 人体の主要な筋が模倣されている.
  Dyneemaの筋ワイヤの末端にはGrommetを使った非線形性要素が配置され, その周りをバネと発泡性素材が囲んでいる(\figref{figure:musculoskeletal-problem}はMusashiの一部の写真である).
  本研究では主に肩3自由度と肘2自由度の5自由度を用いる.
  これらに関係する筋は10本であり, 内二関節筋が一本含まれる.
}%

\begin{figure}[t]
  \centering
  \includegraphics[width=1.0\columnwidth]{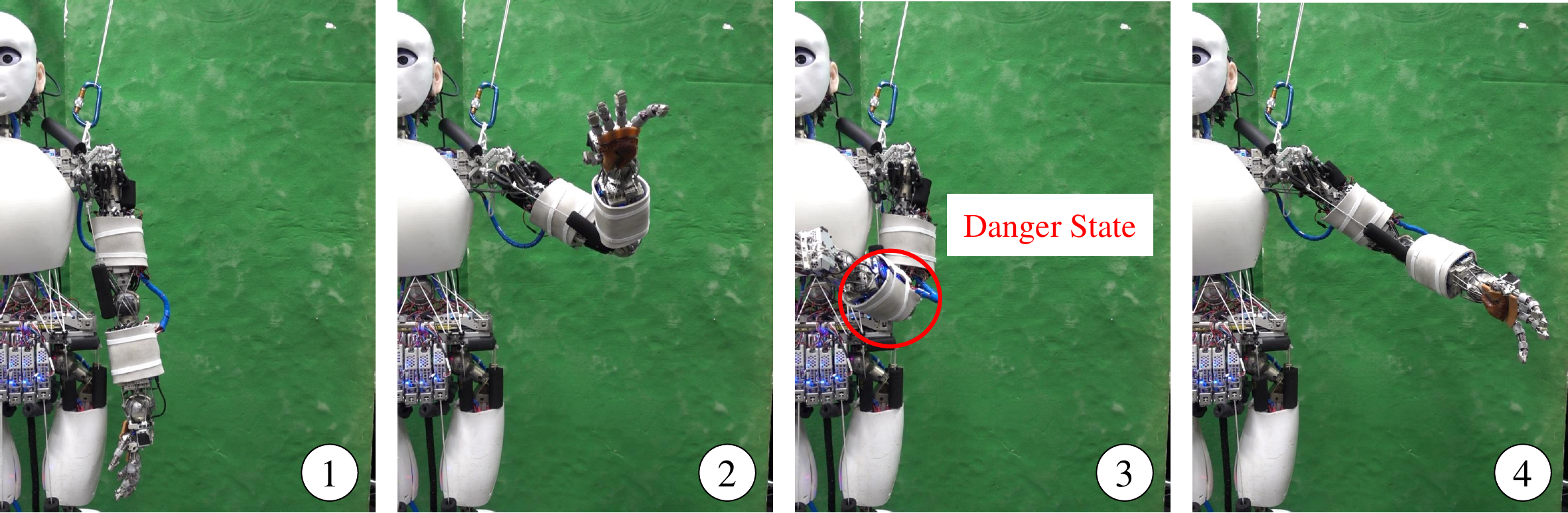}
  \caption{Online learning experiment of Danger Avoidance Network.}
  \label{figure:learning-experiment}
  \vspace{-3.0ex}
\end{figure}

\begin{figure}[t]
  \centering
  \includegraphics[width=1.0\columnwidth]{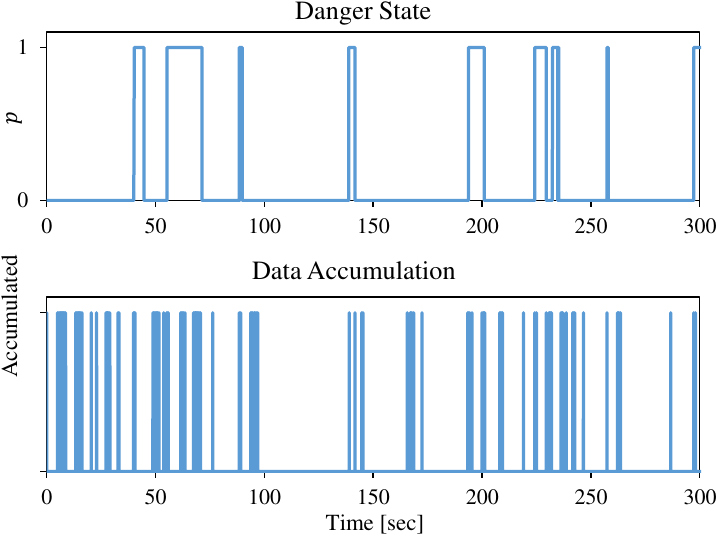}
  \caption{Transition of danger state and data accumulation frequency when conducting online learning.}
  \label{figure:learning-graph}
  \vspace{-3.0ex}
\end{figure}

\begin{figure}[t]
  \centering
  \includegraphics[width=1.0\columnwidth]{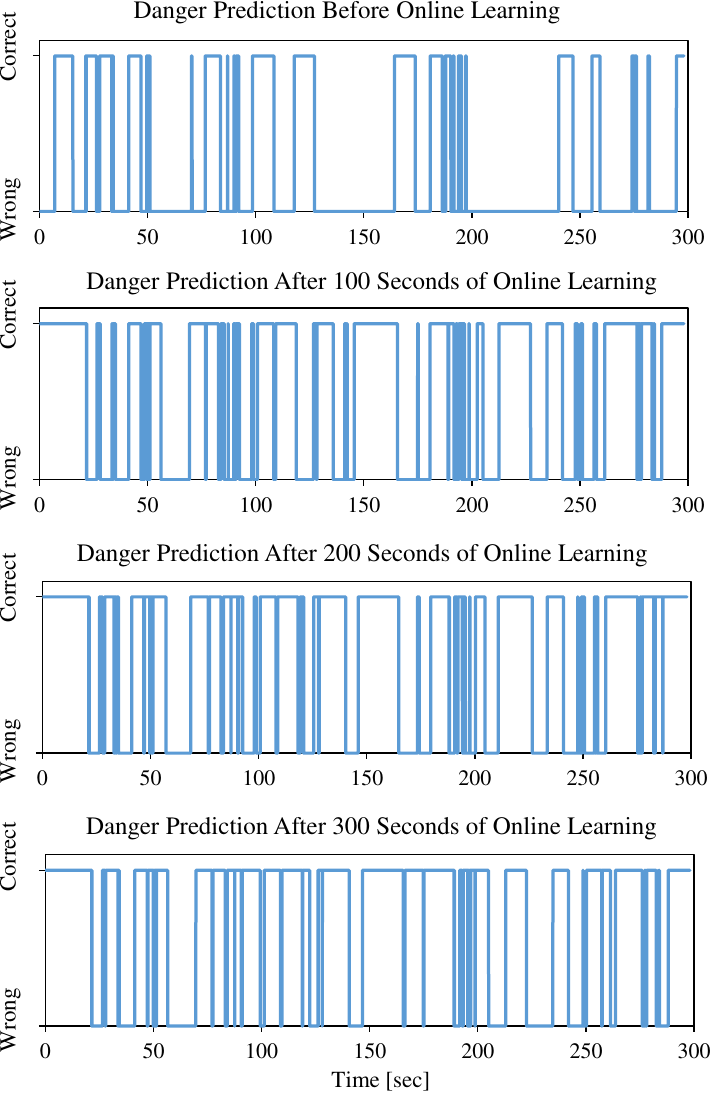}
  \caption{Transition of danger prediction correctness after 0, 100, 200, and 300 seconds of online learning.}
  \label{figure:learning-eval}
  \vspace{-3.0ex}
\end{figure}

\begin{figure}[t]
  \centering
  \includegraphics[width=1.0\columnwidth]{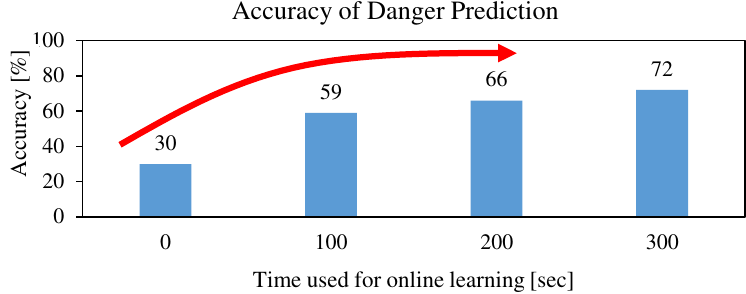}
  \caption{The accuracy of danger prediction after 0, 100, 200, and 300 seconds of online learning.}
  \label{figure:learning-eval2}
  \vspace{-1.0ex}
\end{figure}

\subsection{Experiment of Online Learning} \label{subsec:learning-experiment}
\switchlanguage%
{%
  We performed experiments on online learning of DAN.
  A random joint angle within the lower and upper limits of the joint angle determined by \secref{subsec:initial-training} was converted to a target muscle length by \cite{kawaharazuka2019longtime} with random target muscle tension and sent to the robot as in \figref{figure:learning-experiment}.
  The transition of the danger state and data accumulation frequency, when the safety mechanism and online learning of DAN is executed, is shown in \figref{figure:learning-graph}.
  Danger occurred about 10 times during 300 seconds.
  Data accumulation continued to occur at all times, but it occurred relatively frequently when the state was in danger.

  In order to show the effectiveness of online learning, we performed the same random movements over 300 seconds using the model after 0, 100, 200, and 300 seconds of online learning.
  These random movements are different from those during online learning.
  During this experiment, we predicted $p^{predicted}$ at all times and assumed that $p^{predicted}>0.1$ was the danger state, and the frequency when the predicted danger state was consistent with the danger state calculated from \equref{eq:p-definition} is shown in \figref{figure:learning-eval}.
  It can be seen that the frequency of the conformance gradually increased as the network was updated.
  Here, the calculated probabilities for each case are presented in \figref{figure:learning-eval2}.
  We can see that the probability increased as the learning progresses, reaching 72\% after 300 seconds.
  The probability did not increase significantly afterwards, and oscillated around 70\%.
}%
{%
  DANのオンライン学習に関する実験を行う.
  \secref{subsec:initial-training}で決めた関節角度の下限と上限の範囲内のランダムな関節角度を\cite{kawaharazuka2019longtime}により指令筋長に変換し, \figref{figure:learning-experiment}のようにロボットに送り続ける.
  これと同時に安全機構・DANのオンライン学習を実行させたときのDanger State, データ蓄積の遷移を\figref{figure:learning-graph}に示す.
  300秒間で約10回程度危険が続く状態が起きている.
  データ蓄積は常に起き続けているが, 危険状態のときは比較的頻繁に起きている.

  オンライン学習の効果を示すために, オンライン学習の0, 100, 200, 300秒後のモデルを使って, オンライン学習のときは異なる同様のランダムな動作を300 sec行う.
  このとき, 常に$p^{predicted}$を計算して$p^{predicted}>0.1$の場合を危険状態とし, これが現在の\equref{eq:p-definition}から計算した危険状態と合致していた場合の頻度を\figref{figure:learning-eval}に示す.
  学習するに従って徐々に合致の頻度が増えていくことが読み取れる.
  ここで, それぞれについて合致する確率を計算したものを\figref{figure:learning-eval2}に示す.
  学習が進むごとに確率は上がっていき, 300秒後には 72\%まで上がっていることがわかる.
  この後学習を続けても, 確率が大きく上がることはなく, 70\%付近を振動していた.
}%

\begin{figure}[t]
  \centering
  \includegraphics[width=1.0\columnwidth]{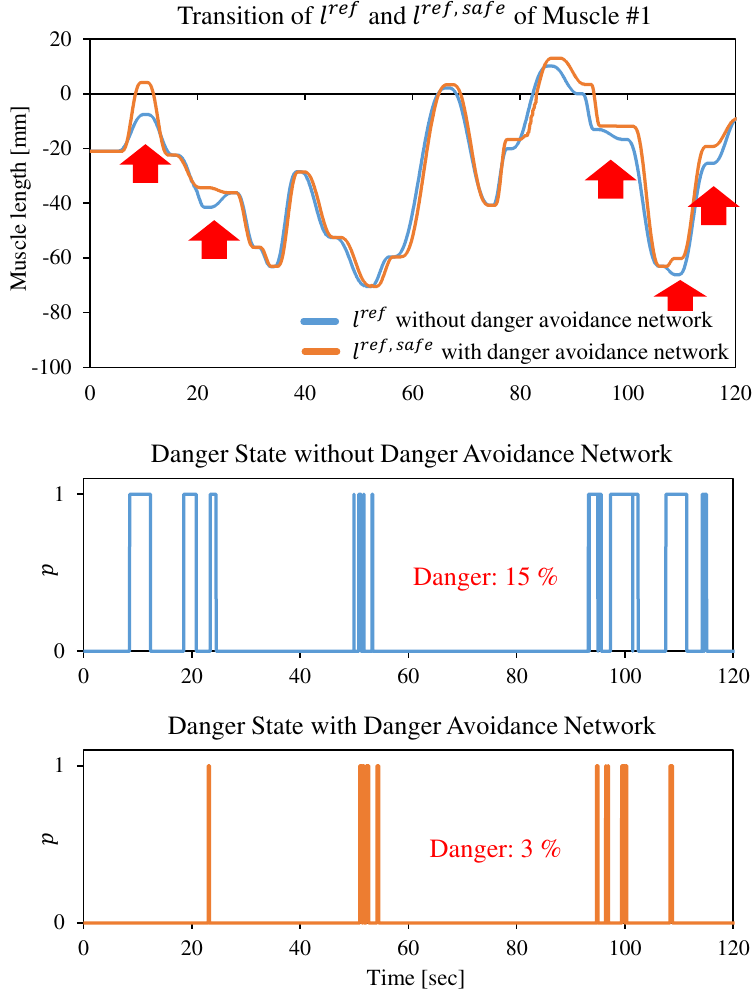}
  \caption{The transition of target muscle length and danger state with and without muscle length modification using danger avoidance network.}
  \label{figure:safe-graph}
  \vspace{-3.0ex}
\end{figure}

\subsection{Experiment of Muscle Length Modification} \label{subsec:safe-experiment}
\switchlanguage%
{%
  We performed experiments on the modification to the safe target muscle length as described in the first half of \secref{subsec:applications}.
  We used the model obtained after 300 seconds of online learning in \secref{subsec:learning-experiment}.
  We moved the robot randomly as in the previous experiment.
  We compared the cases with and without the muscle length modification technique ($\bm{l}^{ref, safe}$ is sent to the actual robot, versus $\bm{l}^{ref}$ is sent to the actual robot) for the same random motion.
  The transition between $l^{ref}$ and $l^{ref, safe}$ of muscle \#1 and the transition of the danger state are shown in \figref{figure:safe-graph}.
  Although there are 10 muscles, only one muscle is shown for better visibility.
  From the transition of muscle length, it can be seen that the muscle length increased in several places by using the muscle length modification, that is, the muscle was elongated.
  If we look at the danger state at these places without the muscle length modification, the danger occurred in almost all of these places.
  On the other hand, with the muscle length modification, it can be seen that none or only a few of the dangers occurred in these places.
  The frequency of the danger was  3\% with and 15\% without the muscle length modification.
  In other words, the danger can be accurately predicted and muscle length can be adjusted to prevent the occurrence of the danger.
  On the other hand, at about 50 seconds, the danger occurred with and without muscle length modification.
  The behavior is considered to not be learned in the experiment of \secref{subsec:learning-experiment}.
}%
{%
  \secref{subsec:applications}の前半で述べた, 安全な筋長指令への修正に関する実験を行う.
  \secref{subsec:learning-experiment}で得られた300秒後のモデルを用いて, 先と同様にランダムに身体を動かす.
  このとき同じランダム動作において, 筋長修正手法を用いない場合($\bm{l}^{ref}$を実機に送る)と用いた場合($\bm{l}^{ref, safe}$を実機に送る)を比較する.
  筋\#1の$l^{ref}$と$l^{ref, safe}$の遷移と, 危険状態の遷移を\figref{figure:safe-graph}に示す.
  筋は10本あるため, 見やすいように一本だけ取り出している.
  指令筋長の推移から, 筋長修正を用いることで, 筋長が増えている, つまり筋を伸ばしている箇所が複数見受けられる.
  それらの該当する箇所における筋長修正を用いない場合の危険状態の推移を見ると, ほとんどの箇所で危険状態が起こっていることがわかる.
  これに対して, 筋長修正を用いた場合は, 該当する箇所の危険状態が全く起こっていない, または少しだけ起こっていることがわかる.
  危険状態の頻度は, 筋長修正を用いない場合は15\%, 筋長修正を用いる場合は3\%となった.
  つまり, 危険状態を正しく予測し, 危険状態が起こらないように筋長を修正できていることがわかる.
  一方, 約50秒付近においては, 筋長修正をしない場合とした場合とで, 同様に危険状態が起こっている.
  これは, \secref{subsec:learning-experiment}の実験において学習し切れなかった動作であることが考えられる.
}%

\begin{figure}[t]
  \centering
  \includegraphics[width=1.0\columnwidth]{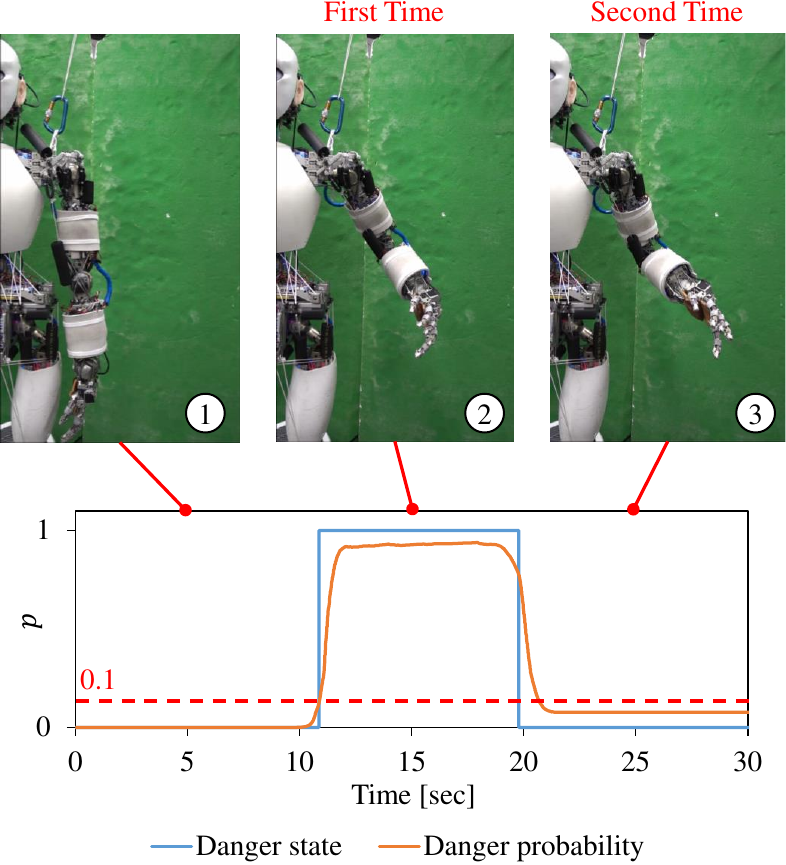}
  \caption{The experiment of prioritized inverse kinematics using DAN. The lower graph shows the transition of $p$ and $p^{predicted}$.}
  \label{figure:ik-experiment}
  \vspace{-3.0ex}
\end{figure}

\subsection{Experiment of Using DAN for Prioritized Inverse Kinematics} \label{subsec:ik-experiment}
\switchlanguage%
{%
  We performed experiments on the prioritized inverse kinematics using DAN as described in the second half of \secref{subsec:applications}.
  In this experiment only, we set $f^{thre}=150$ [N].
  A certain $\bm{x}^{ref}$ was set and the inverse kinematics was solved from a certain initial posture.
  The danger probability was predicted, and if it was dangerous, the inverse kinematics was solved again with this posture as a condition which should be avoided.
  The first and second postures are shown in the upper figure of \figref{figure:ik-experiment}.
  The transition between the predicted danger probability and the actual danger state are shown in the lower figure of \figref{figure:ik-experiment}.
  In the first case, the danger probability was predicted to be 0.96, and the actual state was in danger when the robot was actually moved.
  On the other hand, in the second case, the danger probability was predicted to be 0.08 and the actual state was not in danger when the robot was actually moved.
}%
{%
  \secref{subsec:applications}の後半で述べた, DANを用いた優先度付き逆運動学に関する実験を行う.
  本研究では, $f^{thre}=150$ [N]としている.
  ある$\bm{x}^{ref}$を設定し, ある初期姿勢から逆運動学を解く.
  このときの危険度を予測し, 危険な場合にはその姿勢を条件に入れ, 再度逆運動学を解く.
  この一度目, 二度目の姿勢を\figref{figure:ik-experiment}の上図に示す.
  このときの予測された危険度と, 実際の危険状態の遷移を\figref{figure:ik-experiment}の下図に示す.
  一度目は0.96と危険が予測され, 実際に動かした際にも危険状態が起こっている.
  これに対して, 二度目は危険度が0.08と予測され, 実際に動かした際にも危険状態は起こっていない.
}%

\section{Discussion} \label{sec:discussion}
\switchlanguage%
{%
  We will discuss the results of the experiments.
  We found that the accuracy of the prediction of the danger state increases with the time of learning, while at the same time the accuracy does not increase more than a certain degree.
  This reason is because the prediction does not converge at the borderline between the occurrence and non-occurrence of the danger.
  In addition, it was found that a pinpoint danger can be avoided by using the learned network for muscle length modification.
  Normally, the target muscle length is not changed, but the muscle length is loosened slightly only when a danger is expected, thus preventing the danger.
  Finally, it is found that the task can be performed in a less dangerous posture by using prioritized inverse kinematics.
  As we accumulate dangerous postures, we can search for a new posture that is apart from them.

  There remain some problems in this study.
  First, in the case of whole inter-body interference, we must use $\bm{l}^{ref}$ of all muscles as the input of the network.
  However, the more the number of input states increases, the more difficult the learning becomes and an appropriate grouping of muscles may be required as in \cite{kawaharazuka2018estimator}.
  In addition, it will be necessary to incorporate DAN into the motion planning because the modification to a safe muscle length can change the robot motion.
}%
{%
  実験結果について議論を述べる.
  オンライン学習における実験では, 学習時間が増えるごとに危険状態の予測精度が上がっていくことがわかった.
  同時に, その予測精度はある程度以上は上がらず, 途中で停滞することがわかった.
  これは, 危険が起こるか起こらないかの境目における状態では予測が収束しないことが理由であると考えられる.
  また, 安全な筋長修正においては, 学習したネットワークを用いることで, ピンポイントに危険を回避することができることがわかった.
  通常は指令筋長を変化させず, 危険が予期される場合のみ筋長を少し緩め, 危険状態を未然に防ぐことができる.
  最後に, 優先度付き逆運動学を利用することで, 危険でない姿勢でタスクを遂行できることがわかった.
  危険な姿勢を蓄積していき, それらから離れた姿勢の探索が可能となる.

  本研究の問題点について述べる.
  まず, 全身の干渉を行う場合は全身の筋の$\bm{l}^{ref}$をネットワークの入力としなければならない.
  しかし, 入力の状態数が増えるほど学習は難しくなり, 適切な筋のグルーピングが必要になる可能性がある.
  また, 安全な筋長への修正はロボットの動きを変化させてしまうため, DANを動作計画に組み込んでいくことが今後必要になると考える.
}%

\section{CONCLUSION} \label{sec:conclusion}
\switchlanguage%
{%
  In this study, we described a network configuration and its applications for danger avoidance in the complex body of musculoskeletal humanoids.
  In order to perform danger avoidance in such a complex body, online learning using actual robot sensor information is useful.
  We consider the activation of the safety mechanism according to the muscle tension to be a danger state, and update the network to predict the danger probability from the target muscle length online.
  By using this network, we can modify the target muscle length to a safe one and use DAN to avoid dangerous postures with prioritized inverse kinematics.
  In the future, we would like to use this network for more complex motion planning.
}%
{%
  本研究では, 筋骨格ヒューマノイドの複雑な身体において危険回避を行うためのネットワーク構成とその応用について述べた.
  その複雑な身体において危険回避を行うためには実機データを用いたオンライン学習が有用である.
  筋張力に応じた安全機構が作動したときを危険状態と見なし, 指令筋長から危険度を予測するネットワークを学習させていく.
  このネットワークを用いることで, 危険を回避するように修正された指令筋長を計算したり, 優先度付き逆運動学で危険な姿勢を回避するような方法に用いることができる.
  今後, このネットワークをより複雑な動作計画に利用していきたい.
}%

\section*{Acknowledgement}
This research was partially supported by JST ACT-X Grant Number JPMJAX20A5 and JSPS KAKENHI Grant Number JP19J21672.
The authors would like to thank Yuka Moriya for proofreading this manuscript.

{
  \bibliographystyle{IEEEtran}
  \bibliography{main}
}

\end{document}